\documentclass[letterpaper, 10 pt, conference]{ieeeconf}  

\IEEEoverridecommandlockouts                              

\usepackage{graphicx} 
\usepackage{amsmath} 
\usepackage{amssymb}  
\usepackage{amsbsy}

\usepackage{todonotes}
\usepackage{verbatim}

\usepackage{algorithm}
\usepackage{algpseudocode}

\usepackage[utf8]{inputenc}

\usepackage{cite}
\usepackage{subcaption}
\usepackage{booktabs}
\usepackage{tikz}
\usepackage{pgfplots}

\pgfplotsset{grid style={dotted, gray}}
\pgfplotsset{minor grid style={dotted,gray}}
\pgfplotsset{every tick label/.append style={font=\tiny}}
\pgfplotsset{every axis/.append style={font=\small}}
\pgfplotsset{ylabel near ticks}
\pgfplotsset{xlabel near ticks}
\newlength\figureheight 
\newlength\figurewidth

\title{\LARGE \bf
A geometric approach for learning compliant motions from demonstration
}

\author{Markku Suomalainen and Ville Kyrki
\thanks{This work was supported by Academy of Finland, decision 286580.}
\thanks{M.\ Suomalainen and V.\ Kyrki are with School of Electrical Engineering, Aalto University, Finland {\tt\small markku.suomalainen@aalto.fi, ville.kyrki@aalto.fi}}}

\begin{document}

\maketitle
\thispagestyle{empty}
\pagestyle{empty}

\begin{abstract}
This paper proposes a method to learn from human demonstration compliant contact motions, which take advantage of interaction forces between workpieces to align them, even when contact force may occur from different directions on different instances of reproduction. To manage the uncertainty in unstructured conditions, the motions learned with our method can be reproduced with an impedance controller. Learning from Demonstration is used because the planning of compliant motions in 3-D is computationally intractable. The proposed method will learn an individual compliant motion, many of which can be combined to solve more complex tasks. The method is based on measuring simultaneously the direction of motion and the forces acting on the end-effector. From these measurements we construct a set of constraints for motion directions which, with correct compliance, result in the observed motion. Constraints from multiple demonstrations are projected into a 2-D angular coordinate system where their intersection is determined to find a set of feasible desired directions, of which a single motion direction is chosen. The work is based on the assumption that movement in directions other than the desired direction is caused by interaction forces. Using this assumption, we infer the number of compliant axes and, if required, their directions. Experiments with a KUKA LWR4+ show that our method can successfully reproduce motions which require taking advantage of the environment.
\end{abstract}

\section{INTRODUCTION}
\label{intro}
\begin{figure}[tb]
	\centering
	\includegraphics[width=.75\columnwidth]{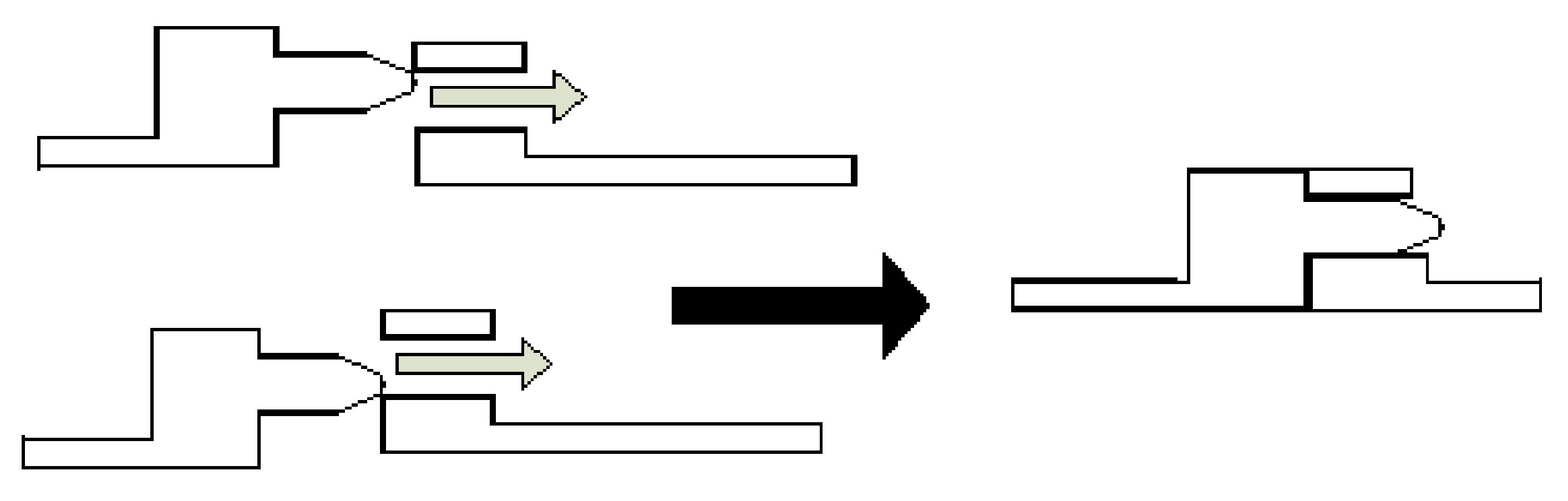}
	\caption{Compliant motions can be used to align workpieces.} 
	\label{fig:assembly}
\vspace{-1em}
\end{figure}

Many assembly tasks require high precision. Humans manage such tasks through taking advantage of interaction forces caused by contacts between objects. These kind of motions are often called compliant motions. This paper looks into the question how compliant motions can be programmed easily. An example of compliant motion in workpiece alignment is depicted in Fig.~\ref{fig:assembly}.  

%

To perform compliant motions, the interaction forces between objects must be limited. The simplest method for this is hybrid force-position control which, however, requires switching if the contact configuration changes and often exhibits force overshoot at contact transitions \cite{Alkkiomaki-icarcv2006}. To alleviate this, it is beneficial to be able to use the same controller for both free space and contact motions. Impedance control is a natural choice for this \cite{hogan1987stable}. An impedance controller features a virtual spring with adjustable stiffness, which allows a controlled amount of deviation from the planned trajectory.

Impedance control has the drawback that it makes motion planning difficult. Preimage planning \cite{lozano1984automatic} can be used for planning compliant motions in two dimensions. However, in 3-D preimage planning has been shown to be computationally infeasible \cite{canny1987new}. There is an on-going interest in developing methods for motion planning under uncertainty \cite{Koval-ijrr2016}.

%

In contrast to automatic planning, learning from demonstration (LfD) \cite{argall2009survey} can be used to transfer skills from a human expert to a robotic system. Most LfD methods try to reproduce any kinds of motions, and use tools such as dynamic movement primitives (DMP) \cite{schaal2006dynamic} or hidden semi-Markov models \cite{racca2016}. Recently, there has been research to use DMP for assembly and workpiece alignment as well \cite{peternel2015human}. However, as DMP's use attractors to form the trajectory, they cannot take full advantage of geometries such as in Fig.~\ref{fig:assembly}, where contact force from either side could be used to guide the workpiece into the correct place. 

 
We propose a method for learning compliant assembly motions from demonstrations. The intuition of our method stems from geometry: there is always a certain range of angles from which a human can push an object to make it slide along a surface. We use kinesthetic teaching to collect the demonstrations. From these demonstrations, both position and force data are recorded. From one or more recordings we learn the direction which will lead the end-effector through the motion, either directly in free-space or in contact. In addition, we learn the necessary compliant axes for performing the current task. 

To allow the recording of forces in kinesthetic teaching, the robot must be equipped with a force/torque sensor and have a gravity compensation mode. The main application of our method is an assembly task consisting of phases and we assume that each phase can be defined by a motion with a single desired motion direction, possibly with directions of compliance to facilitate sliding in contact. Consequently, our method can be viewed as a primitive in a complex task. The combination of primitives for more complex tasks is outside the scope of this paper, but there are numerous publications on the topic (for example \cite{kroemer2014learning}).

The biggest advantage our method has over traditional LfD methods is the aforementioned ability to take advantage of the current task's geometry. With suitable workpiece design, this will greatly increase the the robustness of the method. Force measurement is only required when collecting the demonstrations, and the sensor can be removed for reproduction since there is no force feedback. It is not necessary to know the exact location or orientation of the target workpiece due to the robustness against position errors, and therefore vision is not strictly required. This enables the use of our method in more difficult and unpredictable environments than, for example, DMP for workpiece alignment. 

Our earlier work \cite{suomalainen2016} considered a similar problem under the assumption that the forces exerted by human during demonstration can be directly measured, as is the case for example in teleoperation. In contrast, this work provides a solution to a more general problem where only the interaction forces can be measured.


Section II reviews related work in use of compliant motions and LfD. In Section III we explain the model used for learning the direction of motion and compliant axes in various scenarios. Next, experiments with a KUKA LWR4+ robot arm and their results are presented in Section IV. Finally, the results are discussed and future work is outlined in Section V.

\section{RELATED WORK}
\label{RELATED}
The benefits of impedance control and compliant motions in assembly tasks have been known for a long time \cite{wang1998passive} and are still under active study \cite{stolt2015robotic}. However, most industrial robots sold today are still not equipped to perform impedance control \cite{chen2013robotic}. In industrial settings, passive compliance \cite{yun2008compliant} is often used to limit forces at time of initial contact. In contrast, this paper considers active compliance using impedance control in order to have adjustable compliance. Similar idea of motions constrained by physical restraints has been applied to manipulating articulated objects \cite{petrivc2014online} by learning task dynamics. In contrast, in this paper we aim to identify a control strategy which is applicable over a variety of dynamics.




There are a few approaches for learning compliance parameters.  Kronander and Billard \cite{kronander2012online} proposed learning them from human demonstration by halting a trajectory demonstration at intervals and using direct human demonstration of desired compliance by wiggling the robot. Later they developed a method for also increasing the compliance by gripping the robot tighter \cite{kronander2014stiffness}. However, wiggling a tool which is in contact is not physically feasible and therefore the method is not suitable for in-contact tasks as considered in this paper. In addition, our proposed method estimates the parameters from demonstrations without extra halting.

Automatic learning of the stiffness matrix has been proposed by Carrera et al.~\cite{carrera2015learning} and Rozo et al.~\cite{rozo2013learning}. However, Carrera et al.\ used DMPs for the learning. We observed that DMPs cannot reproduce motions where at a stage of the motion there are multiple correct directions of motion, such as inserting a tool to a funnel. Rozo et al.\ learned the stiffness matrix from positional deviation using least squares estimation in a cooperative assembly scenario. The main difference between our method and Rozo et al.\ is that we require the interaction forces to complete the task, whereas in their work the forces were an unnecessary consequence of the collaborative partner's actions. In our previous publication \cite{suomalainen2016} we assumed that it is possible to directly measure the forces initiated by the human teacher, but in this paper we solve the more general case where only the interaction forces between the tool and the environment can be observed. 

\section{METHOD}
\label{METHOD}

\begin{figure}[tb]
	\centering
	\includegraphics[width=.75\columnwidth]{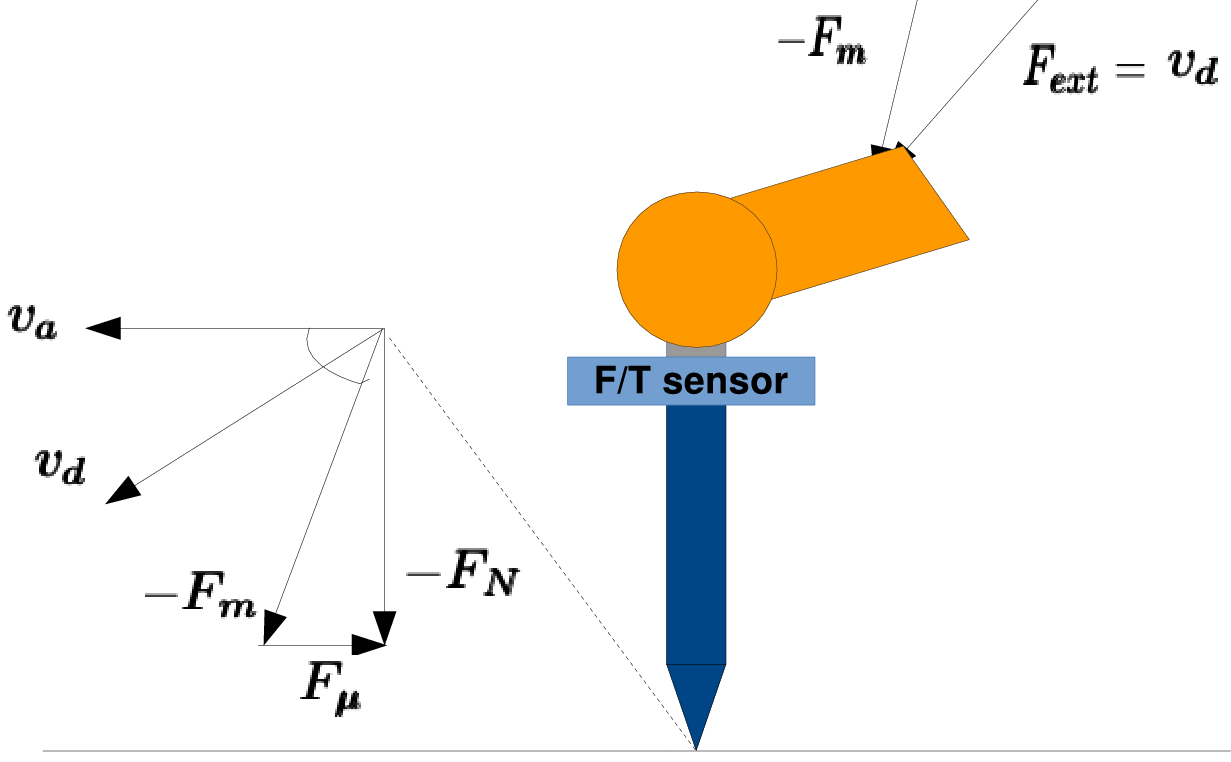}
	\caption{Force/torque sensor configuration, the position where external force by the human teacher $\pmb{F_{ext}}$ is applied and the forces which sum up to the reading of the force/torque sensor.} 
	\label{fig:sliding_forces}
\vspace{-1em}
\end{figure}

We assume that a force/torque sensor is mounted between the location where the human holds the robot and the end-effector. One possible setup is illustrated in Fig.~\ref{fig:sliding_forces}. The force/torque sensor measures the forces between the sensor and the contact point in a Cartesian coordinate system with compensation of the gravity force of the tool. Then when sliding the end-effector along a surface, the force measured by the force/torque sensor $\pmb{F_m}$ is the combination of normal force of the surface $\pmb{F_N}$, Coulomb friction working against the motion, and acceleration of the end-effector $\pmb{a}$,  
\begin{equation}
  \pmb{F_m} = \pmb{F_N} + \vert\mu\pmb{F_N}\rvert\left( -{\pmb{\hat{v}_a}}\right) + m\pmb{a}
  \label{eqt:measured_force}
\end{equation}
where $\pmb{\hat{v}_a}$ is the direction of tool motion. Throughout this paper, we will use the circumflex notation to denote normalized version of a vector. The forces are defined in the world coordinate system and obtained via forward kinematics. We assume that the demonstrations are performed with almost constant speed, in which case we can ignore the acceleration term $m\pmb{a}$.

Assuming the human demonstrator exerts a constant force and the system is in a stationary case where the velocity is constant, a particular motion direction $\pmb{\hat{v}_a}$ can result from any of a set of force directions presented as unit vectors $\pmb{\hat{v}_d}$, which we call the desired directions (see Fig.~\ref{fig:sliding_forces}). The set of desired directions, $\pmb{\hat{v}_d}$, include all directions between $\pmb{-\hat{F}_m}$ and $\pmb{\hat{v}_a}$, because a force applied along any of those would cause sliding in the direction $\pmb{\hat{v}_a}$. 

Before considering how $\pmb{\hat{v}_d}$ can be estimated from demonstrations, we note that we assume that a complex task can be divided into simple motion segments, each of which has a single desired direction of motion. Applying force to end-effector in this desired direction will bring it to the desired position of the current motion, either by free-space motion or sliding along a surface leading towards the position. This paper does not consider the problem of dividing a demonstration into segments and we assume that the segmentation can be performed. 

\subsection{Learning desired direction}
\label{sec:desired}
We define the desired position of a motion as a point or a set of points where the demonstrator wants the end-effector to move to. The desired direction of motion $\pmb{\hat{v}_{d}^*}$, chosen from the set of possible desired directions $\pmb{\hat{v}_d}$, is a direction vector in 3-D Cartesian space. If the robot applies a force in this direction, the end-effector will move towards the desired position, either by free-space motion or by sliding along a surface. The whole process of finding $\pmb{\hat{v}_{d}^*}$ is formulated in Algorithm \ref{algo}.

\begin{figure}[tbp]
\centering
\begin{subfigure}[b]{0.47\columnwidth}
	\centering
	\includegraphics[width=\columnwidth]{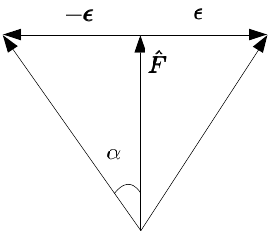}
	\caption{Human variation in 2-D} \label{fig:error_vector}
\end{subfigure}
\begin{subfigure}[b]{0.47\columnwidth}
	\centering
	\includegraphics[width=\columnwidth]{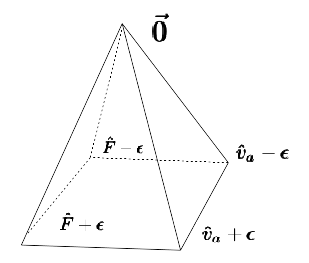}
	\caption{$P$ in 3-D} \label{fig:single_pyramid}
\end{subfigure}
\caption{Visualizations of how (a) the error $\pmb{\epsilon}$ from (\ref{eqt:error}) is added to both $\pmb{\hat{F}}$ and $\pmb{\hat{v}_{a}}$ and (b) how the polyhedron $P$ from (\ref{eqt:polyhedron}) is constructed at each time step.}
\end{figure}

From Fig.~\ref{fig:sliding_forces} and (\ref{eqt:measured_force}) we observe that in the direction of motion, $\pmb{\hat{v}_d}$ is constrained by $\pmb{\hat{v}_{a}}$ and $-\pmb{\hat{F}_m}$. In addition, we consider that a human can not demonstrate a desired trajectory perfectly, resulting in small differences between the demonstrated $\pmb{\hat{v}_{a}}$ and human intent. Because contact constrains the motion, this needs to be taken into account only perpendicular to $\pmb{\hat{v}_{a}}$. We define the maximum error $\pmb{\epsilon}$ such that it forms an angle of $\alpha$ degrees with the plane spanned by unit vectors $\pmb{\hat{v}_{a}}$ and $\pmb{\hat{F}}\equiv-\pmb{\hat{F}_m}$, as illustrated in Fig.~\ref{fig:error_vector}. Formally, we write
\begin{equation}
\pmb{\epsilon} = \tan \alpha \frac{\pmb{\hat{F}} \times \pmb{\hat{v}_{a}} }{\lvert \pmb{\hat{F}} \times \pmb{\hat{v}_{a}} \rvert }
\label{eqt:error}
\end{equation}

Now for each point $\pmb{k}$ along the demonstrated trajectory, we can construct a polyhedron $P$ consisting of the following vectors commencing from origin measured at point $\pmb{k}$:
\begin{equation}
P_k=
\begin{pmatrix}
\pmb{\hat{v}_{a}}+\pmb{\epsilon} \\
\pmb{\hat{v}_{a}}-\pmb{\epsilon} \\
\pmb{\hat{F}}-\pmb{\epsilon} \\
\pmb{\hat{F}}+\pmb{\epsilon} \\
\end{pmatrix}
\label{eqt:polyhedron}
\end{equation}
The polyhedron is illustrated in Fig.~\ref{fig:single_pyramid}. $P$ represents the set of possible desired directions of motion $\pmb{\hat{v}_d}$ of a single point in 3-D. This is described in Algorithm \ref{algo} on row 2.

\begin{algorithm}
\caption{The whole algorithm to calculate desired direction from demonstrations}
\label{algo}
\begin{algorithmic}[1]
\State{Find $R$ which rotates $\pmb{\bar{v}_{a}}$ to positive z axis}
\For {each point $\pmb{k}$} 
\State{Calculate $P$ from (\ref{eqt:error}) and (\ref{eqt:polyhedron})}
\State{Rotate $P$: $P_r = RP$}
\State{Call Algorithm \ref{vec2ang} for each $\pmb{p}$ in $P_r$ to compute $\Theta$} 
\EndFor \\
$G(i,j)=0$ $\forall\; i,j$ 
\For {Each cell $(i,j)$ in G}
\For {Each $\Theta$}
\If {$(i,j)$ inside $\Theta$}
\State{$G(i,j) = G(i,j) + 1$}
\EndIf
\EndFor
\EndFor \\
$g_{i,j}$ indices of vector median of $\max(G)$
\For {each $\Theta$}
\If{$g_{i,j}$ inside $\Theta$}
\State{add $\Theta$ to $\Theta^*$}
\EndIf
\EndFor \\
Compute intersection $\Phi$ of all $\Theta^*$ \\
Calculate Chebyshev center $\pmb{\phi^*}$ of $\Phi$ \\
Call Algorithm \ref{ang2vec} with $\pmb{\phi^*}$ to compute $\pmb{\hat{v}_{des-r}^*}$ \\
Rotate back: $\pmb{\hat{v}_{d}^*} = R^{-1}\pmb{\hat{v}_{des-r}^*}$
\end{algorithmic}
\end{algorithm}


A small value of measured force $\pmb{F_m}$ signifies free space motion and renders the result of normalization in (\ref{eqt:error}) volatile. We detect such cases using a manually chosen threshold for $\pmb{F_m}$. When the force threshold is not met, we construct the polyhedron $P$ from a number of points on a circle with center at $\pmb{\hat{v}_{a}}$, perpendicular to it, and with radius $\pmb{\epsilon}$. With this addition, our method covers pure free space motion as well.

To find the set of feasible desired directions for the whole motion, we must satisfy all the constraints through the trajectory represented by $n$ polyhedra $P$. To simplify the computation, we transform the direction vectors to a 2-D angular space. Thus we project the 3-dimensional polyhedra into 2-dimensional polygons. This is performed by calling Algorithm \ref{vec2ang} with polyhedra $P$.

\begin{algorithm}
\caption{Conversion from Cartesian 3-D vector $\pmb{p}$ to angular 2-D vector $\pmb{\Theta}$.}
\label{vec2ang}
\begin{algorithmic}[1]
\State Normalize $\pmb{p}$
\State $r = \arccos(p_z)$ \\
$\gamma = \arctan2(p_y,p_x)$ \\
$\theta_x = r\cos(\gamma)$ \\
$\theta_y = r\sin(\gamma)$
\end{algorithmic}
\end{algorithm}



To avoid computational problems going from $-\pi$ to $\pi$ radians in the angular coordinate system, we rotate the vectors before conversion such that the resulting angles are centered around zero. To do this, we find a rotation matrix $R$ such that it rotates the mean direction of motion of a demonstration $\pmb{\bar{v}_{a}}$ to match with the positive z axis using Rodrigues formula.  After performing this on row 3 in Algorithm \ref{algo}, we can call Algorithm \ref{vec2ang} with the result. Then we form a rectangle $\Theta$ from the four $(\theta_x$, $\theta_y)$ pairs calculated with Algorithm \ref{vec2ang} describing the constraints in the angle coordinate system, as on row 4 in Algorithm \ref{algo}. These points are the angular coordinates of polyhedron $P$. 

\begin{figure}[tbh]
	\centering
	\includegraphics[width=.65\columnwidth]{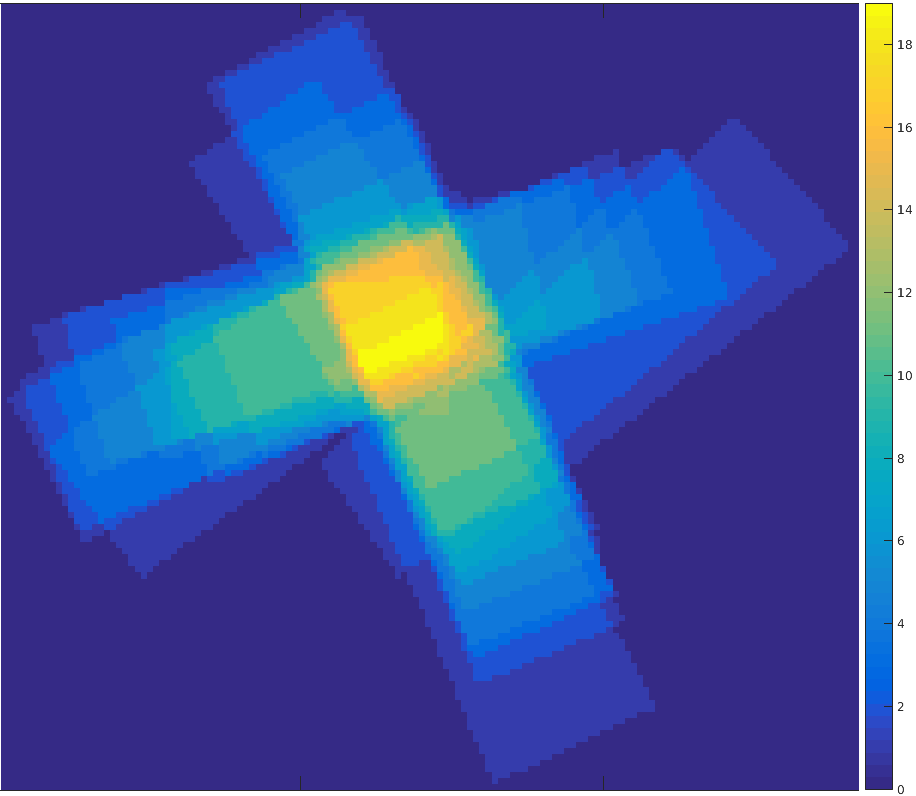}
	\caption{Heatmap illustration of the voting grid depicting the polygon intersection used for outlier rejection. The colormap unit describes the number of $\Theta$ within which each pixel lies.} 
	\label{fig:vote_grid}
\vspace{-1em}
\end{figure}

To find an intersection of these $n$ rectangles, outliers must be rejected. We address this by determining inliers by voting. First we construct a voting grid, matrix $G$ of zeros with a chosen resolution, for example 1 degree, as on row 6 in Algorithm \ref{algo}. The indices of this matrix relate to values in the angular coordinate system. Then for each element in $G$, we check if the element, according to its indices, is inside each polygon $\Theta$. If it is, we add 1 to the value of the element. These are rows 7--13 in Algorithm \ref{algo}. The result is a matrix $G$ with one or more maxima. An example calculated from two perpendicular funnel demonstrations is illustrated in Fig.~\ref{fig:vote_grid}.

If there are more than one maximum values, we choose their vector median \cite{astola1990vector} $g_{ij}$, as described on rows 14--15 in Algorithm \ref{algo}. We then loop over polygons $\Theta$ again and calculate the intersection of the set of polygons $\Theta^*$ in which the indices of $g_{ij}$ lie, as shown on lines 16--20 in Algorithm \ref{algo}. We call the intersection polygon $\Phi$. It describes the set of feasible desired directions $\pmb{\hat{v}_{d}}$ in the angular coordinate system. To find a single direction, $\pmb{\hat{v}_{d}^*}$, we need to choose the corresponding angular system point $\pmb{\phi^*}$ from $\Phi$. We choose $\pmb{\phi^*}$ as the Chebyshev center of $\Phi$, i.e.\ the center of largest circle inscribed in the polygon \cite{garkavi1964chebyshev}. 

Finally, we need to convert $\pmb{\phi^*}$ to a 3-dimensional vector $\pmb{\hat{v}_{des-r}^*}$. This occurs on row 23 in Algorithm \ref{algo} and is presented in detail in Algorithm \ref{ang2vec}. The variable $s$ is needed to deduce the correct sign for the resulting vector, since $\arctan 2$ can produce this information whereas the ratio of $\hat{p_x}$ and $\hat{p_y}$ cannot. The result is a unit vector, but only directions are important in our application.

\begin{algorithm}
\caption{Conversion from angular 2-D vector $\pmb{\Theta}$ to Cartesian 3-D unit vector $\pmb{\hat{p}}$.}
\label{ang2vec}
\begin{algorithmic}[1]
\State{$s = \textrm{sign}(\arctan2(\theta_y,\theta_x))$} \\
$r = \cos\left(\sqrt{\theta_x^2+\theta_y^2}\right)$ \\
$a = \frac{\theta_y}{\theta_x}$ \\
$\hat{p}_z = \arccos(r)$ \\
$\hat{p}_x = s\sqrt{\frac{1-\hat{p}_z^2}{1+a^2}}$ \\
$\hat{p}_y = s\cdot a\cdot \hat{p}_x$
\end{algorithmic}
\end{algorithm}

We still need to rotate $\pmb{\hat{v}_{des-r}^*}$ back to the original coordinate system, by calculating $\pmb{\hat{v}_{d}^*}=R^{-1}\pmb{\hat{v}_{des-r}^*}$. Now the result is the direction along which a force must be applied to guide the end-effector to the desired position. 

If we have multiple demonstrations, the method works exactly the same way. The polygons $\Theta$ are concatenated and an intersection is calculated. Difference between demonstrations simply makes $\Phi$ smaller. If there are multiple viable approach directions, all of them should be demonstrated to allow maximal utilization of environment in reproduction. For example, in the case of the funnel, the demonstrations should come from parallel directions for the method to learn to slide along any side. If two demonstration from nearly same directions are given, the method will assume that this is the only possible direction to approach from and can only replicate such motions. No more than one demonstration from each approach direction is required if the demonstration is well performed. 

\subsection{Learning axes of compliance}
\label{compliance_chapter}
To successfully complete a motion, the compliant axes need to be identified. If the motion occurs only in free space, no compliance is required. In-contact tasks require at least one compliant axis. Certain tasks, such as inserting a tool to a funnel, require a second compliant axis to take full advantage of the funnel's geometry and allow sliding along any side of the funnel. We assume that if compliance is required, the axis should be totally compliant.

We first observe that the compliant axes must be perpendicular to the desired direction $\pmb{\hat{v}_{d}^*}$. This is due to our definition of no stiffness along a compliant axis---the end-effector would not move in that direction even if commanded. The key idea is that if there is movement along other directions besides the desired direction, this movement must be generated by interaction forces. When the environment interaction is causing forces on the end-effector, compliance is needed to reproduce the demonstrated motions. In contrast, if only force but no motion is measured in a certain direction, this direction must be stiff to avoid instability.

\begin{figure}[tbp]
\centering
\begin{subfigure}[b]{0.47\columnwidth}
	\centering
	\includegraphics[width=\columnwidth]{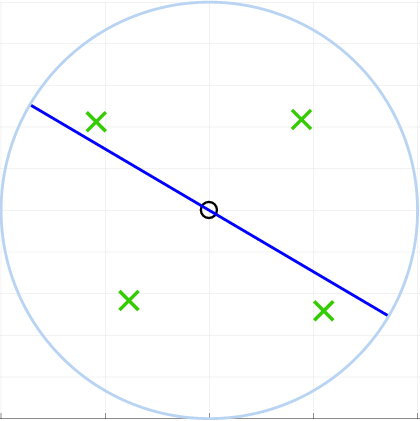}
	\caption{2 compliant axes}
	\label{fig:1_compl}
\end{subfigure}
\begin{subfigure}[b]{0.47\columnwidth}
	\centering
	\includegraphics[width=\columnwidth]{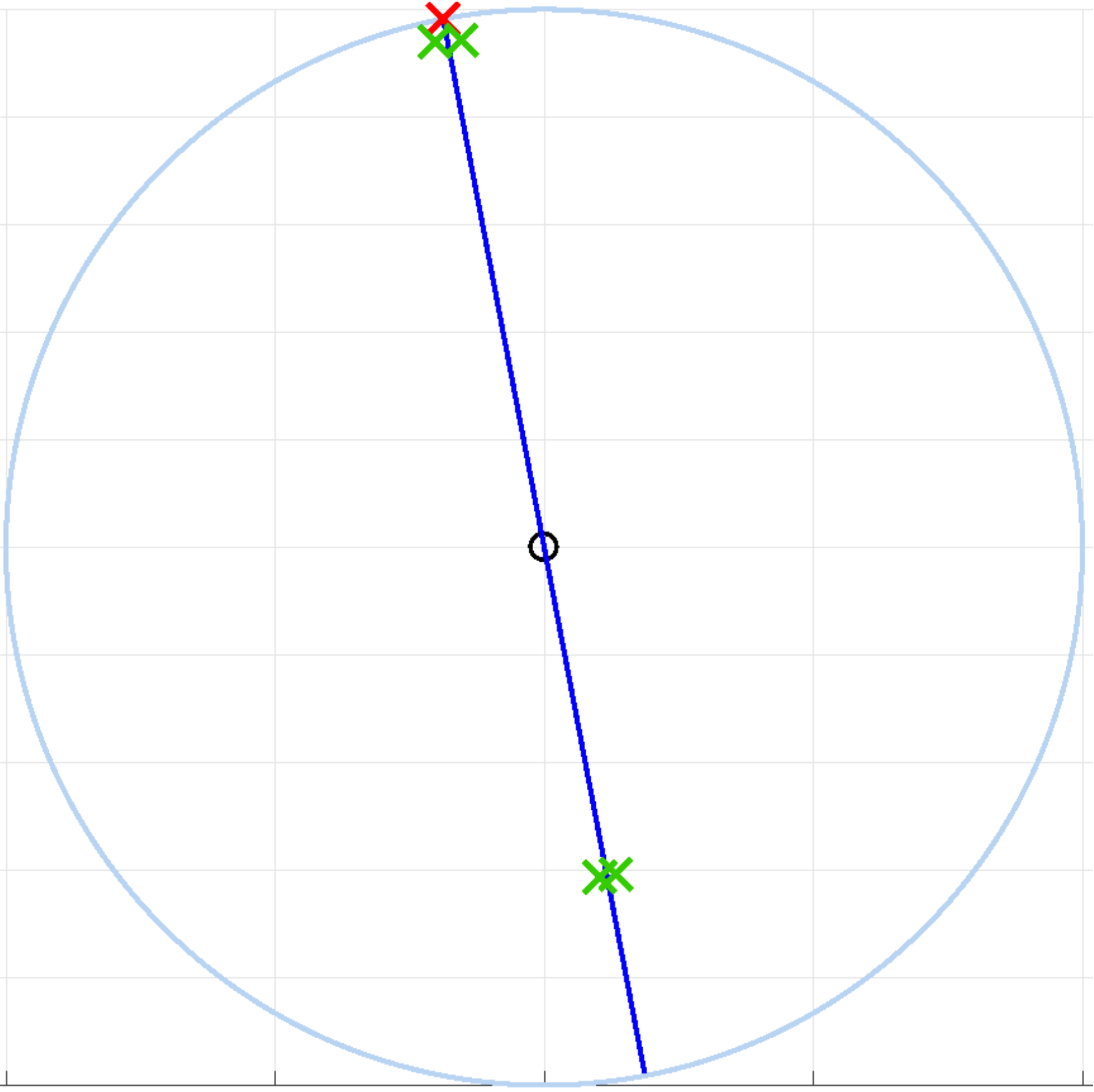}
	\caption{1 compliant axis}
	\label{fig:2_compl}
\end{subfigure}
\caption{Unit spheres in the coordinate system for determining the compliant axes. Origin (small black circle) represents $\pmb{\hat{v}_{d}^*}$ in the angular coordinates, green crosses are the corresponding $\pmb{\phi_a}$ of each demonstration, blue line is the linear model $u$ and red cross the identified compliant direction when the model with 1 compliant axis is chosen.}
\label{fig:comp_axes}
\end{figure}

To exploit this idea, we first calculate the mean direction of actual motion $\pmb{\bar{v}_a}$ separately for each demonstration used in calculation of corresponding $\pmb{\hat{v}_{d}^*}$. Then we rotate all $\pmb{\bar{v}_a}$ such that in the new coordinate system $\pmb{\hat{v}_{d}^*}$ is along the z-axis, after which we use Algorithm \ref{vec2ang} to convert the set of $\pmb{\bar{v}_a}$ to angular coordinates $\pmb{\phi_a}$. As a result, the values $\pmb{\phi_a}$ will be within a unit circle, where origin represents $\pmb{\hat{v}_{d}^*}$ and the values on the unit circle are directions perpendicular to $\pmb{\hat{v}_{d}^*}$. As we assume compliance is required in the direction where we observe motion without human initiative, we can see that the direction of compliance should be towards $\pmb{\phi_a}$ and lie on the unit circle to fulfil the orthogonality requirement. This is illustrated in Fig.~\ref{fig:comp_axes}.

To choose the correct number of compliant axes, we need to measure how well each model (number of compliant axes) explains the observations, but also discourage the choice of an overly complicated model. We take our inspiration from the Bayesian Information Criterion (BIC) \cite{schwarz1978estimating}, which is defined
\begin{equation}
BIC = \ln(n)k-2\ln(L)
\label{eqt:bic}
\end{equation}
where $n$ is the number of data points, $k$ the number of parameters and $L$ the likelihood of a model. 

We assume that the error a human makes while demonstrating a task is normally distributed with variance $\sigma$. Therefore we calculate the likelihoods for each model from a 2-D normal distribution
\begin{equation}
L_i = \prod_j \mathcal{N}\left( \epsilon_{i,j}\; \vert \; \pmb{0},\begin{pmatrix}
\sigma & 0 \\
0 & \sigma
\end{pmatrix} \right).
\label{eqt:normal}
\end{equation}
where $\epsilon_{i,j}$ is the error of  data point $j$ in model $i$. 

For the case of no compliant axes (i.e. free space motion) under no errors $\pmb{\phi_a}=\pmb{0}$. Therefore, the error is defined $\pmb{\epsilon_0}=\pmb{\phi_a}$. With a single compliant axis, the error is the distance from $\pmb{\phi_a}$ to a line $\pmb{u}$, illustrated in Fig.~\ref{fig:comp_axes}, which maximizes the likelihood of $\pmb{\phi_a}$ and passes through the origin. Because normal distribution is isotropic, we can set $\pmb{\epsilon_1}=[\pmb{u^T}\pmb{\phi_a}\ 0]$. Finally, 2 compliant axes perfectly explain any data as the model describes the linear combination of two lines, and therefore we set $\pmb{\epsilon_2}=[0\ 0]$.

We calculate the BIC for each of the models and choose the one with lowest BIC. If the model with a single compliant axis is chosen, we choose the direction of compliance as the intersection between $\pmb{u}$ and the unit circle, marked with red x in Fig \ref{fig:comp_axes}. If no compliance is needed, a position controller is sufficient. If two compliant axes are needed, any pair of vectors which form an orthonormal base with 
$\pmb{\hat{v}_{d}^*}$ are adequate as the compliant axes. The whole process is summarized in Algorithm \ref{compliant}.

It should be noted that the proposed approach does not follow the typical use of BIC which is only applicable when $n\gg k$ and the variance in the likelihood is calculated from the data. Instead, we assume that the uncertainty of demonstrations can be estimated beforehand, making it possible to use the proposed formulation. 

\begin{algorithm}
\caption{Finding the required number of compliant axes and their directions}
\label{compliant}
\begin{algorithmic}[1]
\State {Find $R$ such that $R\pmb{\hat{v}_{d}}=\pmb{\hat{z}}$}
\State {Calculate set of mean actual directions $\pmb{\bar{v}_a}$}
\State {Rotate $\pmb{\bar{v}_{ar}} = R \pmb{\bar{v}_a}$}
\State {Apply Algorithm \ref{vec2ang} to convert $\pmb{\bar{v}_{ar}}$ to angular coordinates $\pmb{\phi_a}$}
\State {Calculate $L_0=\mathcal{N}\left(\pmb{\phi_a}\vert \pmb{0},\Sigma\right)$}
\State {Calculate $L_1=\mathcal{N}\left(\pmb{\epsilon_1}\vert \pmb{0},\Sigma\right)$, where $\epsilon_1 = [\pmb{u^T}\pmb{\phi_a}\ 0]$}
\State {Calculate $L_2=\mathcal{N}\left(\pmb{0}\; \vert \pmb{0},\Sigma\right)$}
\State {Calculate BIC (\ref{eqt:bic}) of $L_0$, $L_1$ and $L_2$. Choose the model with lowest BIC value.}

\end{algorithmic}
\end{algorithm}

\subsection{Reproduction}

An impedance controller is used for the reproduction. It is a feedback controller defined as
\begin{equation}
  \pmb{F} = K(\pmb{x}^*-\pmb{x})+D\pmb{v}+\pmb{f_{dyn}},
  \label{eqt:imp_control}
\end{equation}
where $\pmb{x}^*$ is the desired position, $\pmb{x}$ the current position, $K$ a gain matrix, $D\pmb{v}$ a linear damping term and $\pmb{f_{dyn}}$ the feed-forward dynamics of the robot including gravity. The reproduction is performed similarly as in \cite{suomalainen2016}. The stiffness is set according to $K=RVR^T$, where $V$ is a diagonal matrix defining the stiffness and the number of compliant axes. $R$ defines the directions of compliant axes and is constructed from $\pmb{\hat{v}_{d}^*}$ and the compliant axes discovered with Algorithm \ref{compliant}. The desired positions for each time step are calculated in a feed-forward manner from $\pmb{\hat{v}_{d}^*}$ and the desired velocity $v$ as
\begin{equation}
  \pmb{x}^*_t= \pmb{x}^*_{t-1}+\pmb{\hat{v}_{d}^*} v  \Delta t
\end{equation}
where $\Delta t$ the sample time of the control loop.

\section{EXPERIMENTS AND RESULTS}
\label{EXPERIMENTS}

The robot used for the experiments was a KUKA LWR4+ lightweight arm. The control law for the Cartesian impedance control of the KUKA LWR4+ through KUKA Fast Research Interface (FRI) \cite{schreiber10} is
\begin{equation}
  \pmb{\tau}=J^T(\mathrm{diag}(\pmb{k_{FRI}})(\pmb{x^*}-\pmb{x})+\mathrm{diag}(\pmb{d_{FRI}})\pmb{v}+F_{FRI})+\pmb{f_{dyn}}
  \label{eqt:kuka}
\end{equation}
where $\mathrm{diag}(\pmb{k_{FRI}})$ is a diagonal matrix constructed of the gain values of $\pmb{k_{FRI}}$. As in \cite{suomalainen2016}, we implemented our controller through the $\pmb{F_{FRI}}$ by setting $\pmb{k_{FRI}}=0$ and $F_{FRI}=K(\pmb{x}^*-\pmb{x})$, getting a controller equal to Eq.~(\ref{eqt:imp_control}) where $ \pmb{f_{dyn}}$ is managed by the KUKA's internal controller. We set the damping value through $\pmb{d_{FRI}}$ to $0.7\frac{N\cdot s}{m}$, the stiffness of the compliant axes through our own controller to 0 and the non-compliant axes to 500 $\frac{N}{m}$ for safety reasons.

\begin{figure}[tbp]
\centering
\begin{subfigure}[b]{0.37\columnwidth}
	\centering
	\includegraphics[width=\columnwidth]{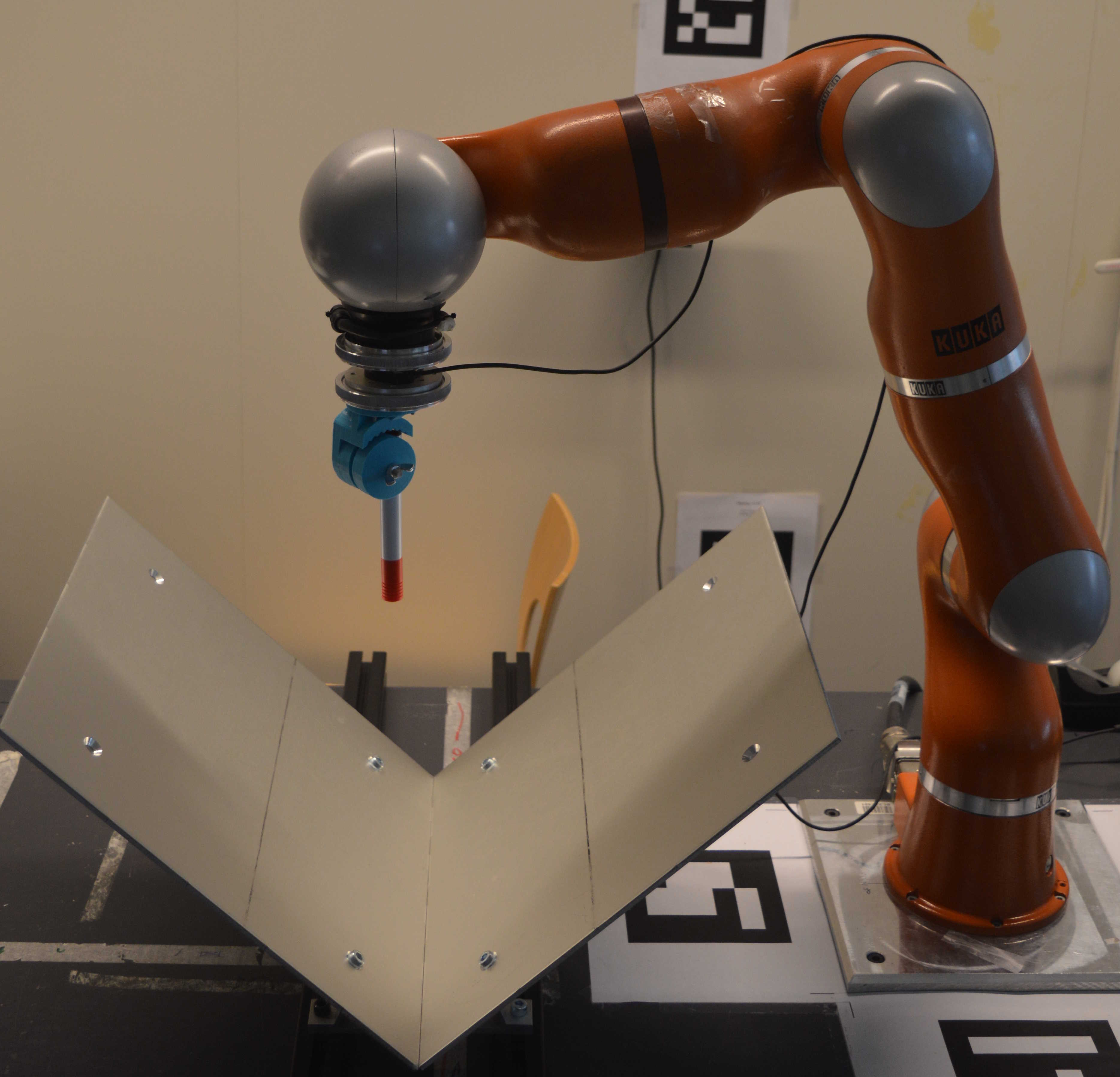}
	\caption{Robot and valley}
\end{subfigure}
\begin{subfigure}[b]{0.47\columnwidth}
	\centering
	\includegraphics[width=\columnwidth]{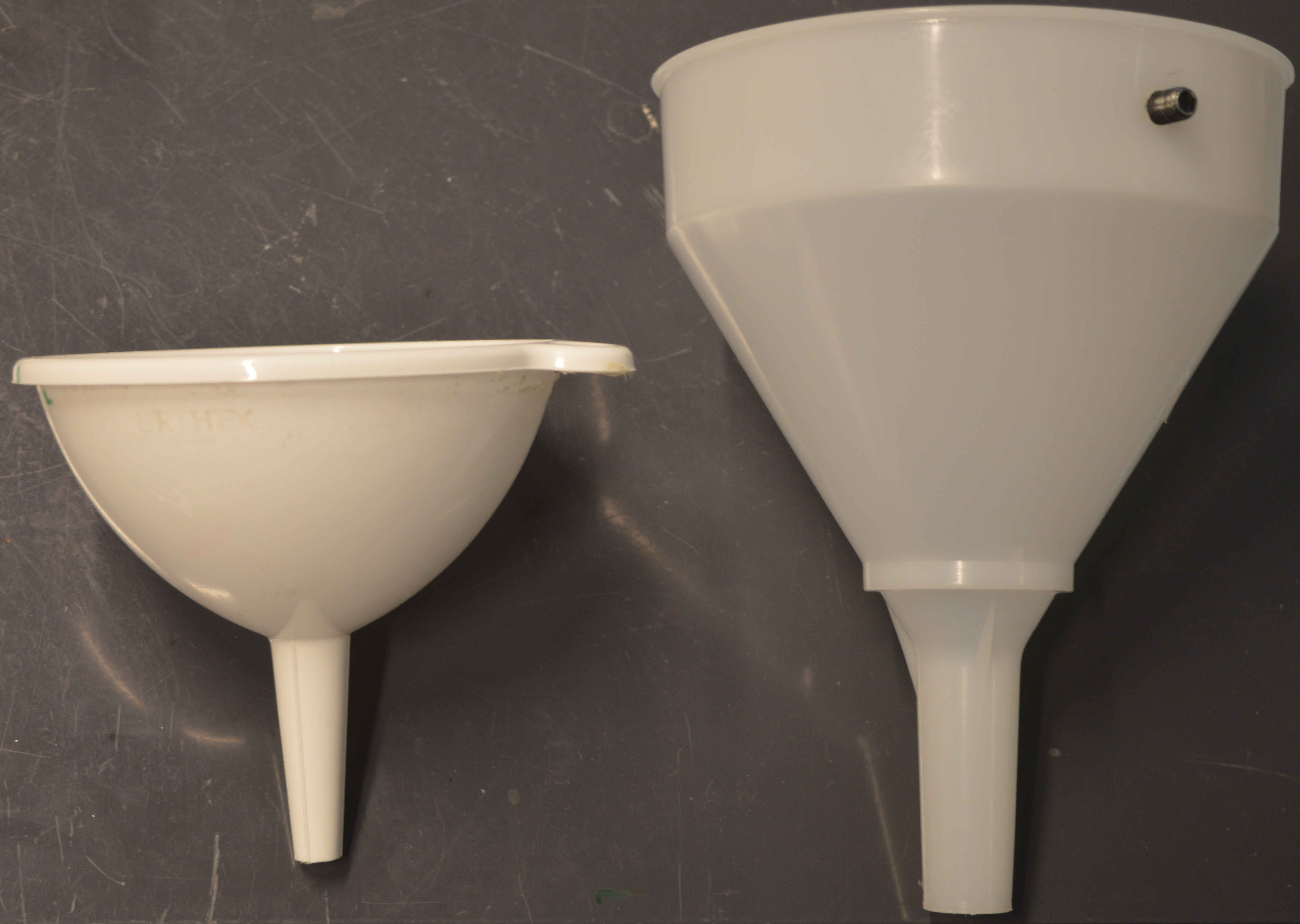}
	\caption{Funnels}
\end{subfigure}
\caption{Our test equipment, the KUKA LWR4+ lightweight arm, the valley setup, and the two funnels, curved and straight.}
\label{fig:setup}
\end{figure}

To test our method, we built two different test setups. First one is a valley consisting of two aluminium plates placed in 90 degrees angle with each other. The second one is a plastic funnel with curved slopes. In addition, we tested the reproduction on another funnel, which had straight slopes. All of the hardware we used is presented in Fig.~\ref{fig:setup}.

In practice, due to noise in the demonstration from human and measurement uncertainty, averaging over a chosen number of time steps to compute $P$ of (\ref{eqt:polyhedron}) produces more stable results. To filter the noise, we chose to average over 20 time steps of original 100Hz measuring frequency, which meant sampling $P$ in 5Hz. We used a manually estimated maximum error angle of 20 degrees for human demonstration for value of $\alpha$ in (\ref{eqt:error}).


\subsection{Desired motion direction}

To validate that we can learn the desired direction from both curved and straight surfaces and multiple directions, we executed a number of experiments on the physical setups presented in Fig.~\ref{fig:setup}. We also wanted to study the number of demonstrations required for learning $\pmb{\hat{v}_{d}^*}$.  Unfortunately, no alternative methods exist in the literature that could be used for comparison, the closest alternative being our earlier work \cite{suomalainen2016} which assumes direct measurement of demonstrator forces. 

\begin{figure}[tbh]
	\centering
	\includegraphics[width=.7\columnwidth]{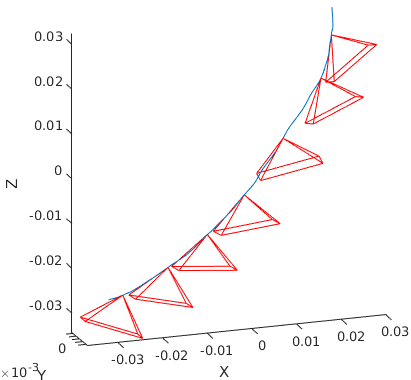}
	\caption{The polyhedra $P$ drawn to a trajectory on a funnel demonstration. Units are in $m$.} 
	\label{fig:traj_rects}
\vspace{-1em}
\end{figure}

To better illustrate the process, we drew the polyhedra $P$ from (\ref{eqt:polyhedron}) at intervals in 3-D in Fig.~\ref{fig:traj_rects}. The trajectory is from a funnel setup, which explains the curved shape. The form of Fig.~\ref{fig:single_pyramid} is distinguishable and it can be seen how the direction of $P$ changes over time. 

\begin{figure}[tbp]
\centering
\begin{subfigure}[b]{0.47\columnwidth}
	\centering
	\includegraphics[width=\columnwidth]{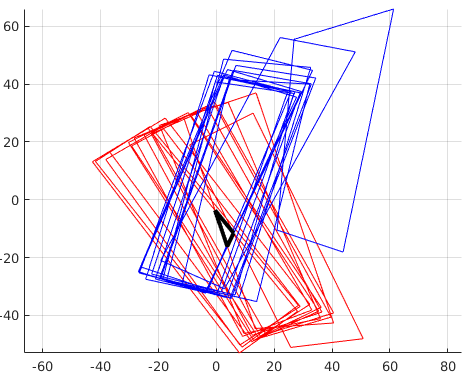}
	\caption{Valley setup} \label{fig:sivusuorat_2d}
\end{subfigure}
\begin{subfigure}[b]{0.47\columnwidth}
	\centering
	\includegraphics[width=\columnwidth]{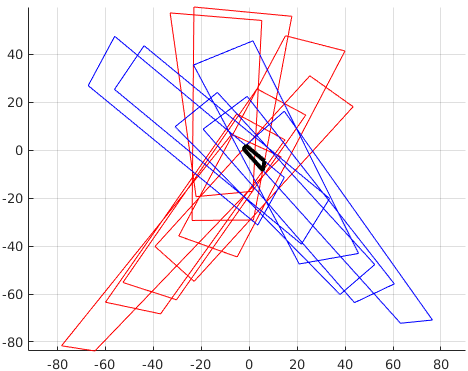}
	\caption{Funnel setup} \label{fig:suppilot_2d}
\end{subfigure}
\caption{2-D polygons $\Theta$ (the axes representing $\theta_x$ and $\theta_y$) of two demonstrations: (a) down different sides of the valley and (b) perpendicularly into the funnel. The red and blue rectangles represent the $\Theta$ of separate demonstrations, and the black polygon is the set of desired directions in angular coordinate system, $\Phi$. }
\end{figure}

The intersection calculations are performed in 2-D on the polygons $\Theta$ calculated in Algorithm \ref{vec2ang} and concatenated from multiple demonstrations. Figure \ref{fig:sivusuorat_2d} shows two trajectory demonstrations performed by sliding along different sides of the valley. The difference in the orientation of the polygons is due to the normal forces being in opposite directions, therefore constraining the final intersection $\Phi$. Figure \ref{fig:suppilot_2d} shows two perpendicular funnel demonstrations. Now we can see that $\Theta$ are almost perpendicular. The funnel demonstrations were briefer than the valley, which explains that there are less $\Theta$ in Fig.~\ref{fig:suppilot_2d}, but still the algorithm finds a clear intersection $\Phi$.

\begin{figure}[tb]
	\centering
	\includegraphics[width=.7\columnwidth]{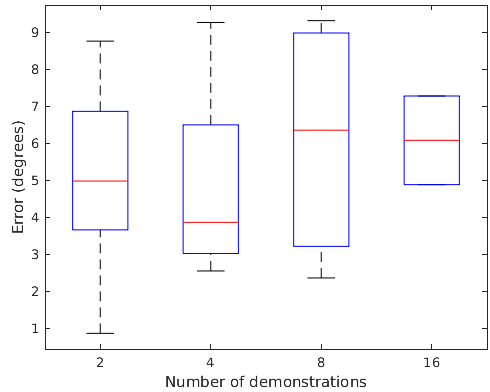}
	\caption{Box plot of the error angle with different number of demonstrations used to calculate $\pmb{\hat{v}_{d}^*}$. The edges of the blue boxes are the 25th and 75th percentiles of the data.}
	\label{fig:suppilot_box}
\vspace{-1em}
\end{figure}

We also wanted to verify our assumption that one demonstration from each viable direction is enough to calculate $\pmb{\hat{v}_{d}^*}$. To do this, we performed 32 demonstrations of the funnel motion, such that all the demonstrations were either perpendicular or opposite to each other. In this case every 4th demonstration came from the same direction. We assumed that in the funnel demonstration, the ground truth of $\pmb{\hat{v}_{d}^*}$ is directly downwards when the funnel is upright. Then we divided the demonstrations into groups of 2, 4, 8 and 16 demonstrations, calculated the $\pmb{\hat{v}_{d}^*}$ and it's error. Box plot of the results is in Fig.~\ref{fig:suppilot_box}. We can see that the error was below our assumption of human error already on 2 demonstrations and did not decrease when more demonstrations were added. Therefore we conclude that one demonstration along each possible trajectory is enough to learn a valid $\pmb{\hat{v}_{d}^*}$.

\subsection{Degrees of freedom}

\begin{figure}[tbp]
\centering
\begin{subfigure}[b]{\columnwidth}
	\includegraphics[scale=0.3]{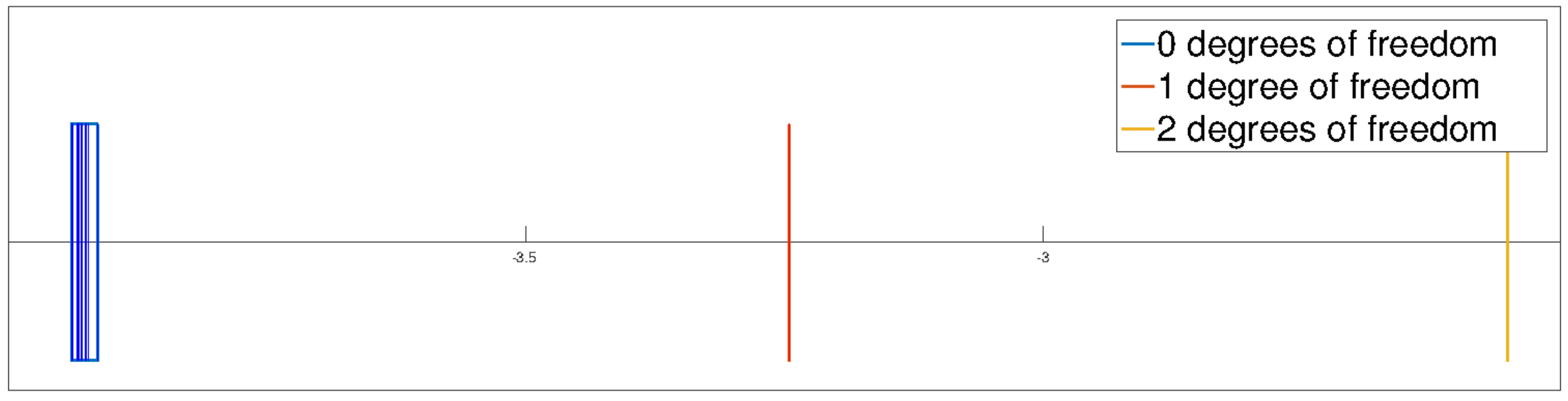}
	\caption{BIC values for free space motion}
	\label{fig:bicfreespace}
\end{subfigure}
\begin{subfigure}[b]{\columnwidth}
	\includegraphics[scale=0.3]{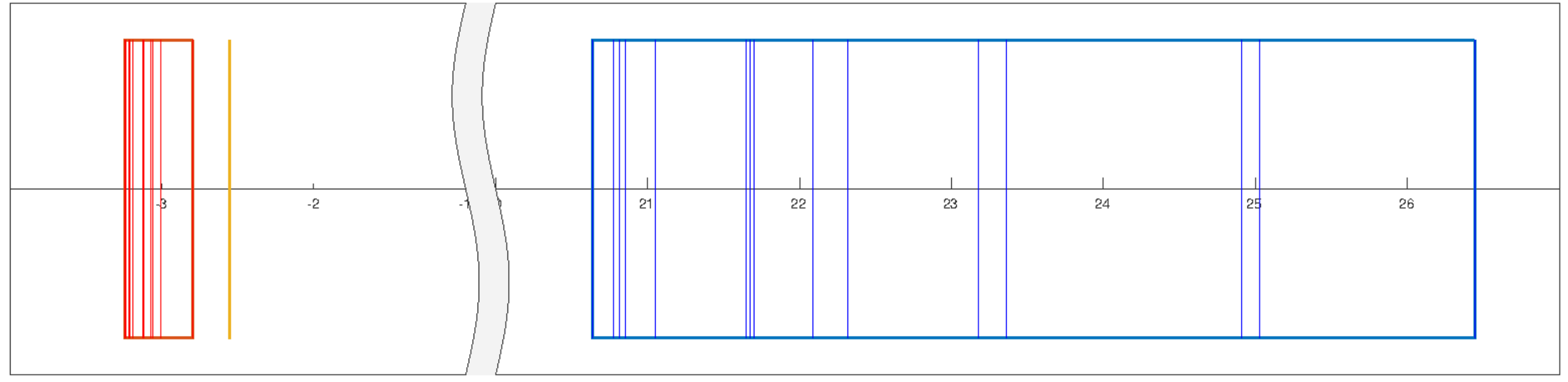}
	\caption{BIC values for the task of sliding to the bottom of the valley}
	\label{fig:bicsuorat}
\end{subfigure}
\begin{subfigure}[b]{\columnwidth}
	\includegraphics[scale=0.3]{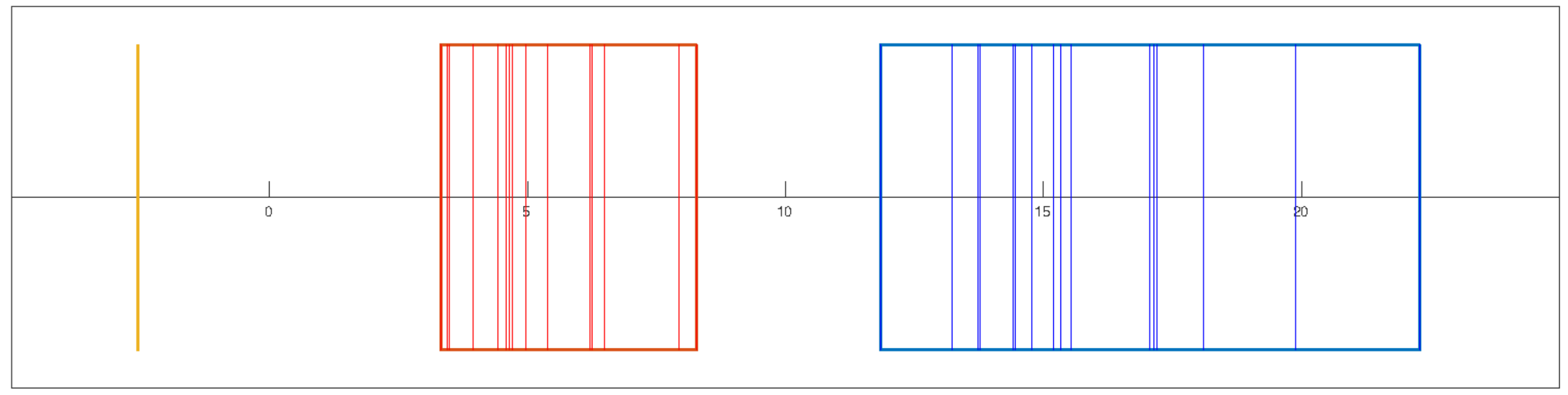}
	\caption{BIC values for insertion to funnel}
	\label{fig:bicsuppilo}
\end{subfigure}
\caption{BIC values for three different experiments}
\end{figure}

To verify our method for finding the degrees of freedom, we performed 30 demonstrations of free space motion with a straight downward trajectory and 30 demonstrations of sliding down the valley. In free space motion zero compliant degrees of freedom are required. Sliding down the valley requires one compliant degree of freedom, since sliding is required but only in a single direction. The funnel demonstration must be reproducible from anywhere within the projection of the funnel, and therefore two compliant degrees of freedom are required. The resulting BIC values can be seen in Fig.~\ref{fig:bicfreespace} for free space motion, Fig.~\ref{fig:bicsuorat} for the demonstrations down the valley and Fig.~\ref{fig:bicsuppilo} for the funnel. We can see that our algorithm performs the classification correctly. The choice of $\sigma$ in (\ref{eqt:normal}) is important for these results. We used the value $\sigma=0.03$, which corresponds to standard error deviation of approximately 10 degrees in the human demonstration.

\begin{figure}[tbp]
\centering
\begin{subfigure}[b]{0.47\columnwidth}
	\centering
	\includegraphics[width=\columnwidth]{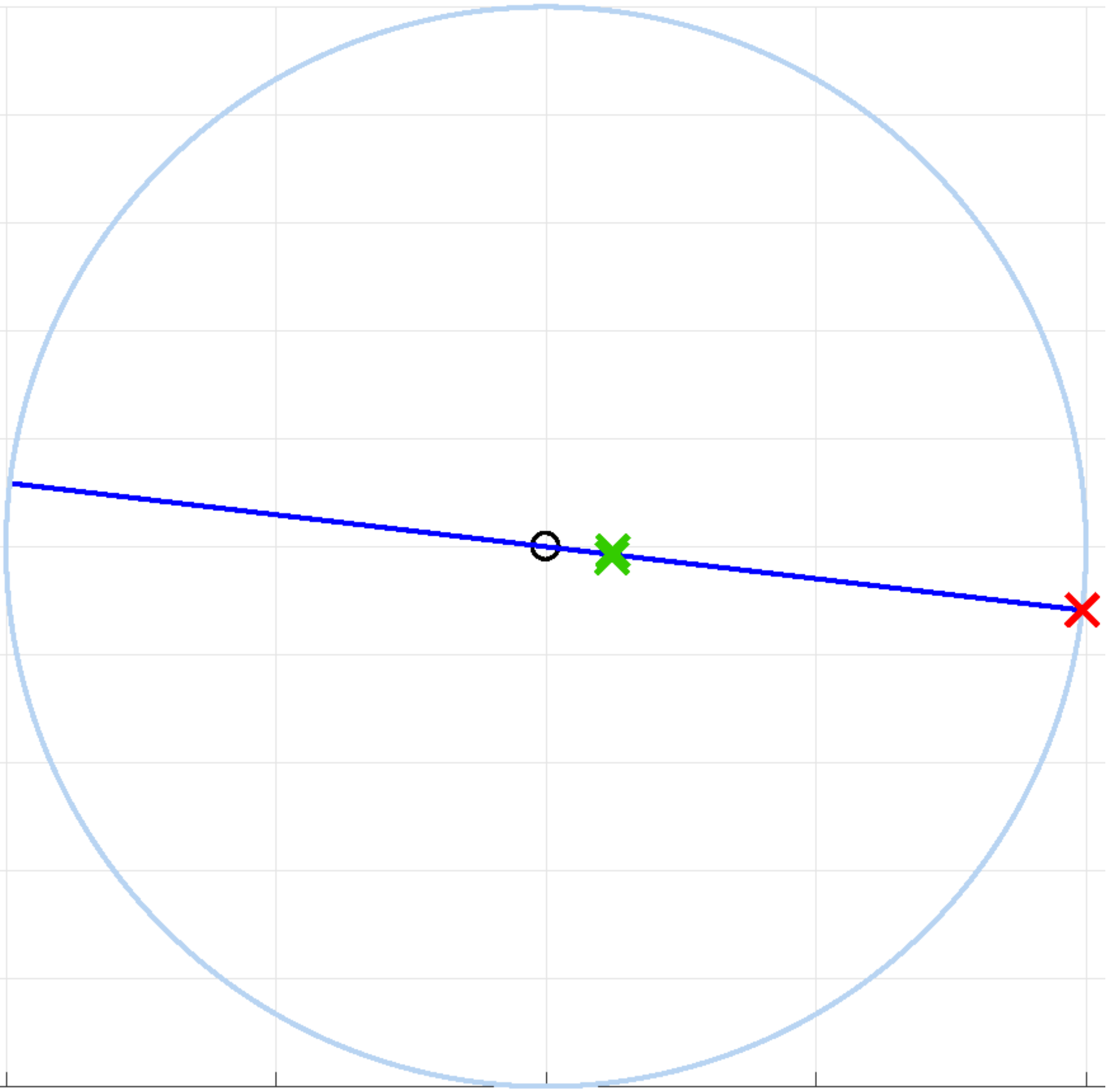}
	\caption{Valid demonstration}
	\label{fig:good_demo}
\end{subfigure}
\begin{subfigure}[b]{0.47\columnwidth}
	\centering
	\includegraphics[width=\columnwidth]{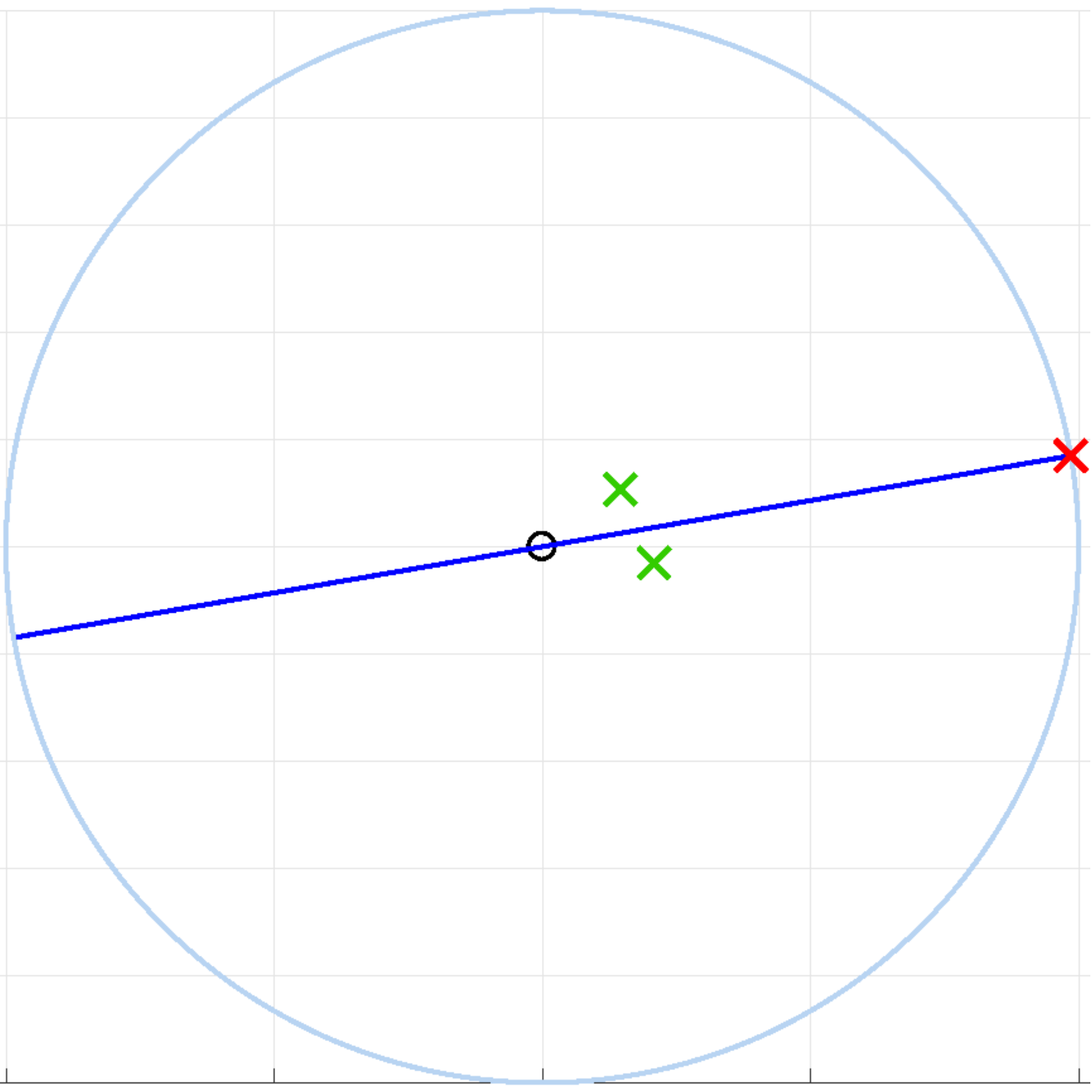}
	\caption{Poor demonstration}
	\label{fig:bad_demo}
\end{subfigure}
\caption{Two demonstrations of sliding along the side of the valley depicted on each figure. Same coordinate system as in Fig. \ref{fig:comp_axes}.} 
\label{fig:sivu_units}
\end{figure}

To study the learning of a challenging motion, we performed 30 demonstrations of sliding along the side of the valley. Unlike the down the valley motion, which in our setup would work even with 2 degrees of freedom, strictly one degree of freedom aligned vertically was required for a successful reproduction. Figure~\ref{fig:sivu_units} shows two combinations of two demonstrations in the same coordinate system as in Fig.~\ref{fig:comp_axes}. In Fig.~\ref{fig:good_demo} the actual directions $\pmb{v_a}$ of the demonstrations are well aligned and the model with 1 compliant axis is correctly identified. However in Fig.~\ref{fig:bad_demo} the demonstrations are not well aligned even if the human demonstrator attempted to demonstrate the same direction both times and therefore our algorithm detects that the model with 2 compliant axes better explains the data. Therefore care must be taken when the demonstrations are given, especially if the minimum number of demonstrations is used for the algorithm and the motion is difficult for the human teacher to demonstrate.

\subsection{Reproduction}
\begin{figure*}[tbp]
\centering
\begin{subfigure}[b]{0.47\columnwidth}
	\centering
	\includegraphics[width=.75\columnwidth]{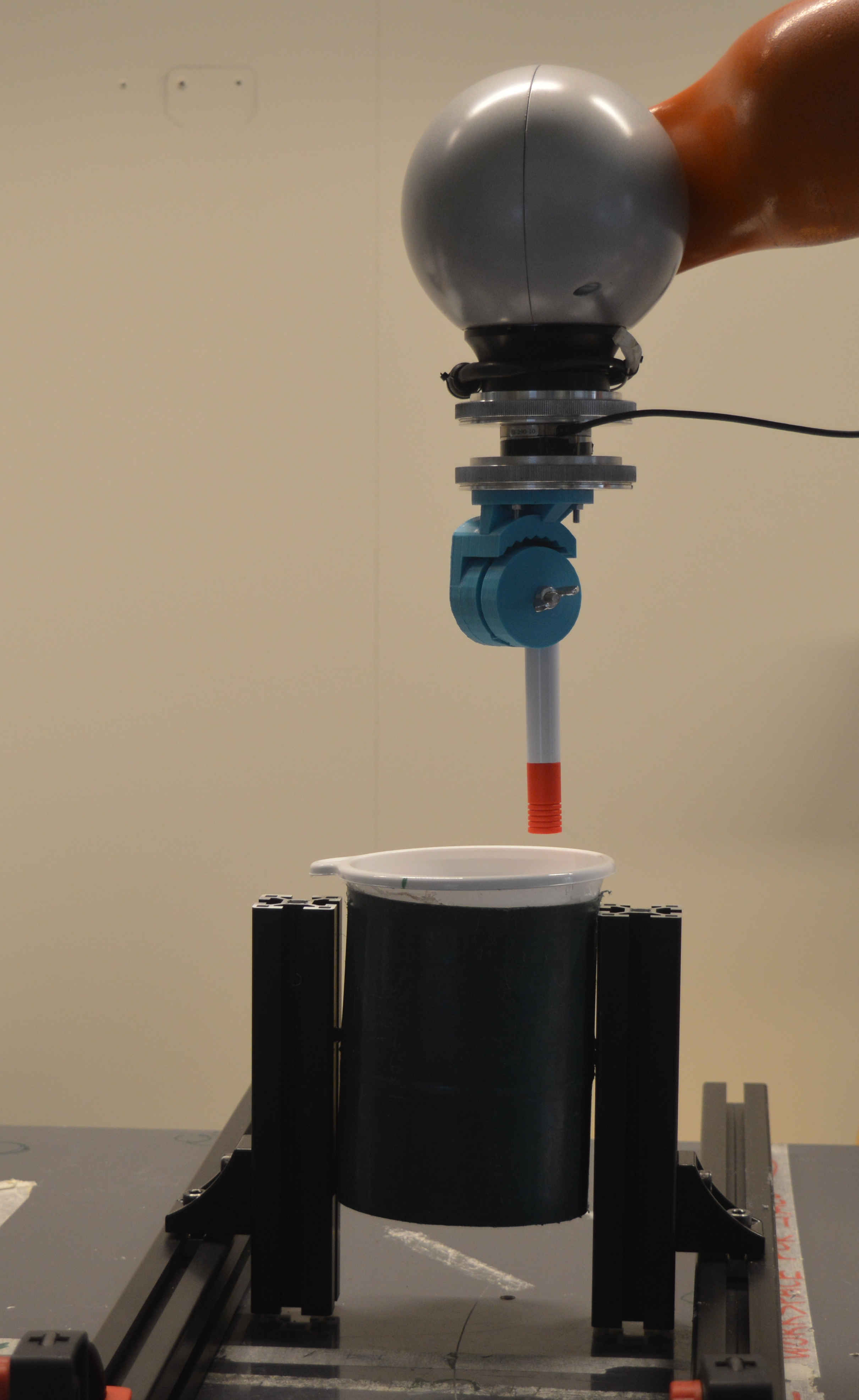}
	\caption{}
	\label{fig:sup1}
\end{subfigure}
\begin{subfigure}[b]{0.47\columnwidth}
	\centering
	\includegraphics[width=.75\columnwidth]{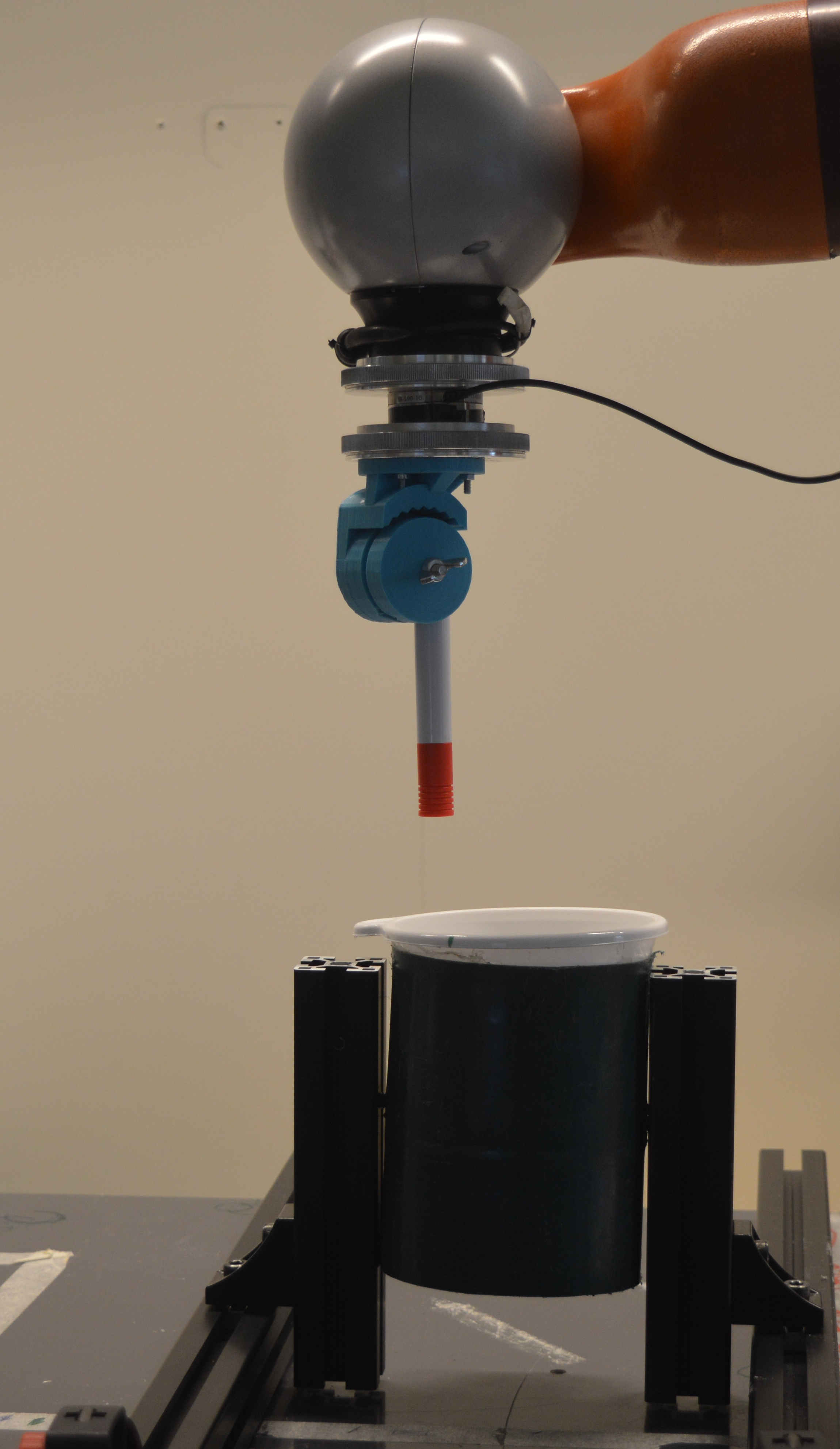}
	\caption{}
	\label{fig:sup2}
\end{subfigure}
\begin{subfigure}[b]{0.47\columnwidth}
	\centering
	\includegraphics[width=.75\columnwidth]{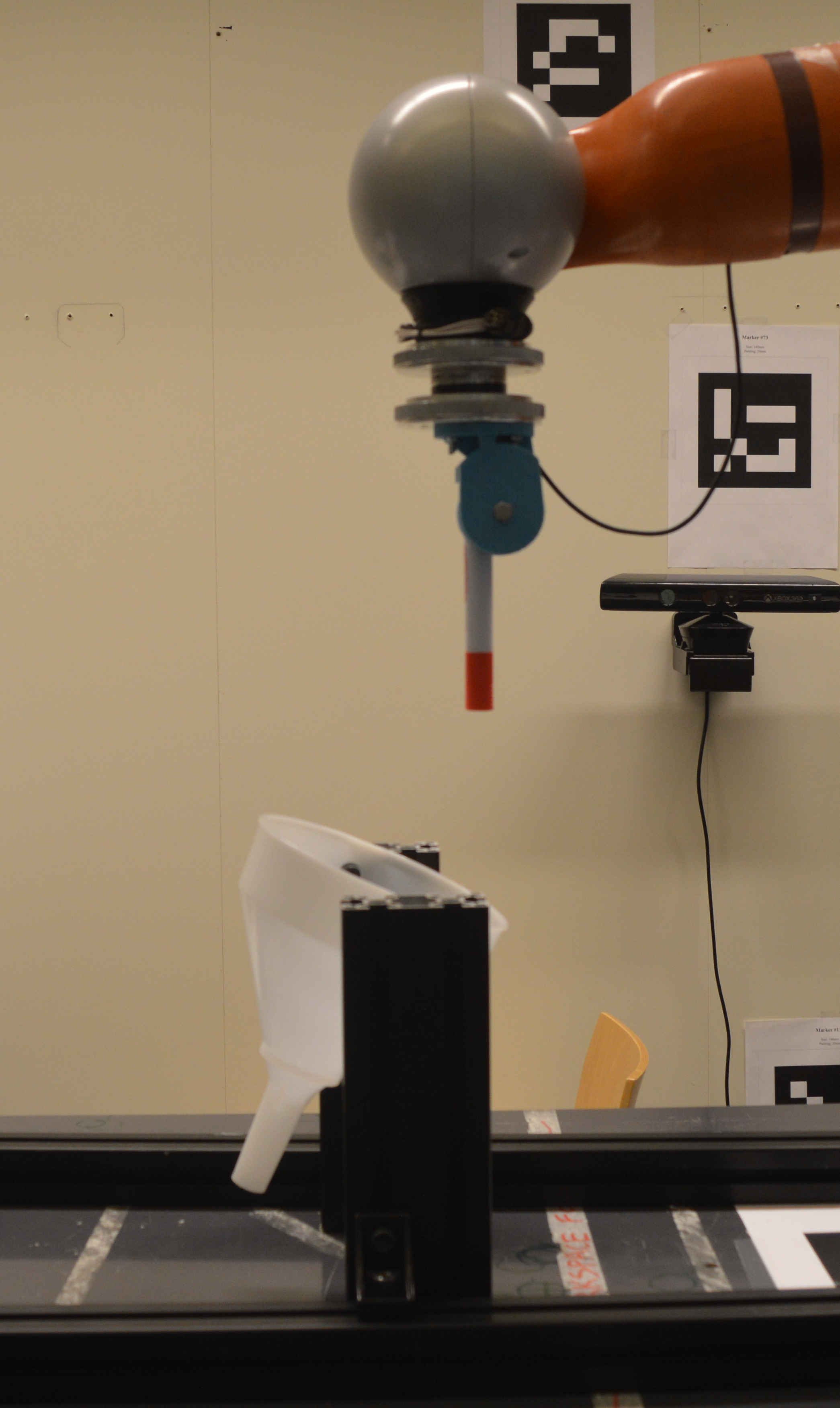}
	\caption{}
	\label{fig:vino1}
\end{subfigure}
\begin{subfigure}[b]{0.47\columnwidth}
	\centering
	\includegraphics[width=.75\columnwidth]{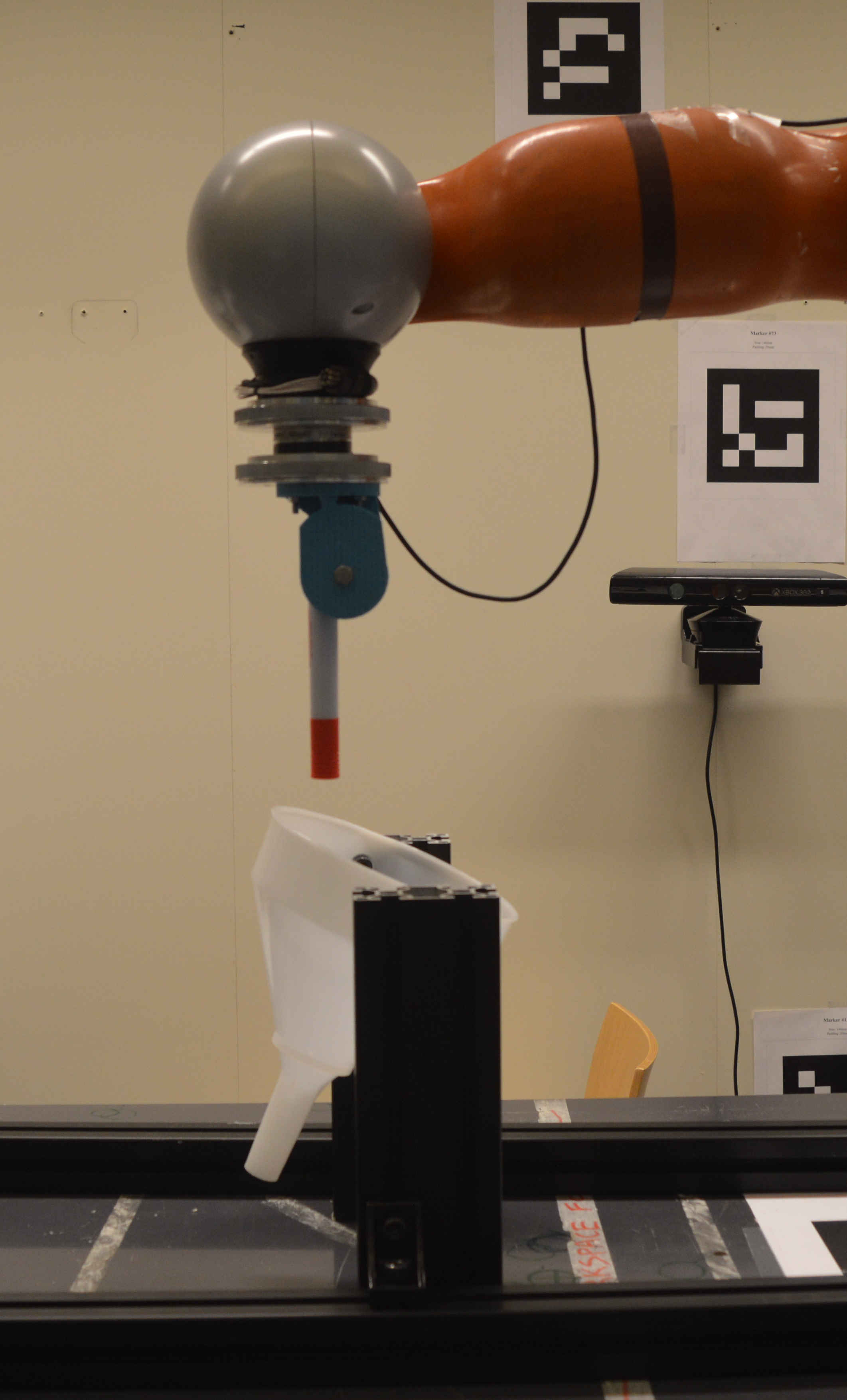}
	\caption{}
	\label{fig:vino2}
\end{subfigure}
\caption{Possible setups in which the end-effector is successfully moved to the target at the bottom of the funnel.}
\label{fig:starting_pos}
\end{figure*}

We studied the execution robustness of our method by varying the starting positions of the motion and also by making changes in the test setups. First our algorithm learned $\pmb{\hat{v}_{d}^*}$  and the model with two compliant axes using 2 perpendicular demonstrations with the curved funnel. Figure~\ref{fig:starting_pos} presents four different starting positions and setups in which the motion was successfully reproduced with the learned parameters. In Figs. \ref{fig:sup1} and \ref{fig:sup2} we used the same funnel as for learning with a varying starting position. In Figs. \ref{fig:vino1} and \ref{fig:vino2} we used the funnel with direct sides and tilted it 15 degrees. In addition, we performed successful reproductions in both valley setups after two demonstrations, sliding down the valley and along the side of the valley. We conclude that our method can successfully learn the required parameters from a small number of demonstrations to perform motions which take advantage of the environment.

\section{CONCLUSIONS AND FUTURE WORK}
We showed that our method can successfully learn to replicate assembly motions which require interaction with the environment. A user will need to give one or more kinesthetic demonstrations of the task using a robot with a force/torque sensor attached between the end-effector and the user's grasp. The motion is fully described by the desired direction, number of compliant axes and their directions. The desired direction is learned using the geometric intuition that a sliding motion is executed with a sufficient force in any direction between the direction of motion and the sum of normal force and Coulomb friction. The compliant axes are learned by assuming that they must be perpendicular to the desired direction and that all motion in other directions besides desired direction must be caused by compliance.

The main advantage of the method compared to existing LfD methods is that the method can learn how to take advantage of physical guides to align workpieces. Since most assembly tasks can be represented as a sequence of motions, learning a single direction together with the compliant axes is sufficient. However, for tasks where contact force modulation is required, other methods are more suitable. 


Our method does not solve a whole assembly task. The demonstration of a whole task must first be divided into motion segments which can be learned with our method. The segmentation could be performed using threshold values \cite{stolt2015robotic} or machine learning methods such as Hidden Markov Models \cite{kroemer2014learning}. The suitability of various segmentation approaches with our method remains a future study.


Many assembly tasks require rotational motions such as screwing. Therefore a natural extension for the method would be to learn rotational motions from demonstration. This has, to our knowledge, not been done before and we will pursue it in our future research.


\label{CONCLUSION}


\bibliographystyle{ieeetr}

\end{document}